# A Remote Sensing Image Change Detection Method Integrating Layer Exchange and Channel-Spatial Differences

Sijun Dong[1], Fangcheng Zuo[1], Geng Chen[2], Siming Fu[1], Xiaoliang Meng,[1,3*]

**Affiliations**

1. Sijun Dong, Fangcheng Zuo, Siming Fu, Xiaoliang Meng* School of Remote Sensing and Information Engineering, Wuhan University, Wuhan, China.
2. Geng Chen Guangxi Water& Power Design Institute CO., Ltd.Minzhu road 1-5, Nanning, Guangxi, 530027, China.
3. Xiaoliang Meng Hubei LuoJia Laboratory, Wuhan University, 430079 Wuhan, China.
*Address correspondence to: xmeng@whu.edu.cn

**Abstract**

Change detection in remote sensing imagery is a critical technique for Earth observation, primarily focusing on pixel-level segmentation of change regions between bi-temporal images. The essence of pixel-level change detection lies in determining whether corresponding pixels in bi-temporal images have changed. In deep learning, the spatial and channel dimensions of feature maps represent different information from the original images. In this study, we found that in change detection tasks, difference information can be computed not only from the spatial dimension of bi-temporal features but also from the channel dimension. Therefore, we designed the Channel-Spatial Difference Weighting (CSDW) module as an aggregation-distribution mechanism for bi-temporal features in change detection. This module enhances the sensitivity of the change detection model to difference features. Additionally, bi-temporal images share the same geographic location and exhibit strong inter-image correlations. To construct the correlation between bi-temporal images, we designed a decoding structure based on the Layer-Exchange (LE) method to enhance the interaction of bi-temporal features. Comprehensive experiments on the CLCD, PX-CLCD, LEVIR-CD, and S2Looking datasets demonstrate that the proposed LENet model significantly improves change detection performance. The code and pre-trained models will be available at: https://github.com/dyzy41/lenet.

# INTRODUCTION

Change detection in remote sensing imagery stands as a cornerstone in the realm of Earth observation. By comparing images of the same geographic location acquired at different times, this process identifies and quantifies changes on the Earth's surface. It is extensively applied in domains such as environmental monitoring, urban planning, natural disaster assessment, and land use change detection, serving as a vital tool for understanding and managing dynamic Earth processes. However, conventional machine learning-based change detection methods, which depend on manually engineered features and rules, often struggle to capture the complexities and variations of surface environments [1].

The advent of deep learning technology in recent years has unlocked new possibilities and achieved remarkable progress in remote sensing change detection [2]. Deep learning models, particularly Convolutional Neural Networks (CNNs) [3] and Transformers [4], with their powerful feature extraction and pattern recognition capabilities, can imgA　　　　imgA　　　　label　　　layer1　　　layer2　　　layer3　　　layer4　　similarity automatically learn effective representations from large amounts of data, significantly improving the accuracy and efficiency of change detection tasks. As computational resources expand and large-scale remote sensing datasets grow, the integration of deep learning into Earth observation has emerged as a prominent research focus [5]. This paradigm not only advances the performance of change detection tasks but also drives the evolution of intelligent Earth observation systems.

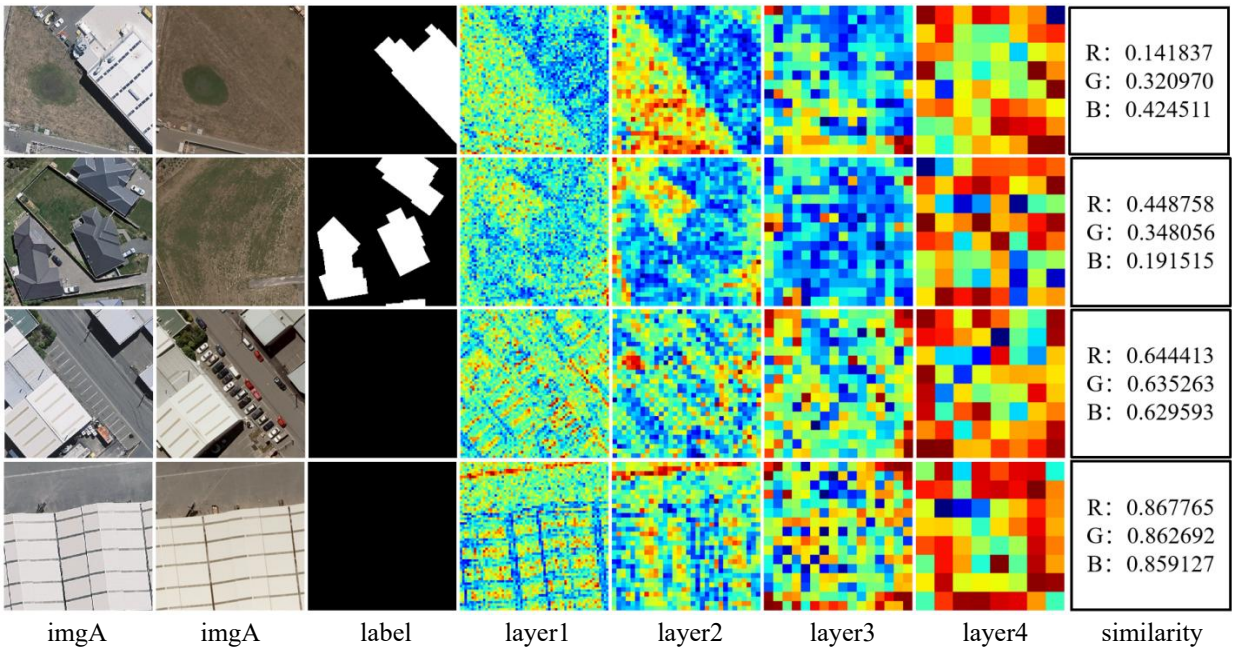

*Fig. 1. The manifestation of differences across various dimensions in change detection imagery. The calculation method for Layer1 to Layer4: The Swin Transformer V2 is used to encode the images, and the cosine similarity in the spatial dimension is computed separately. The calculation method for Similarity: The bi-temporal images are flattened according to the RGB channels, and the cosine similarity between the two vectors is calculated.*

In deep learning-based pattern recognition tasks, feature extraction serves as a cornerstone of success. Researchers continually refine feature extraction models to drive advancements in various downstream tasks [6], [7], [8]. Improvements in feature extraction architectures have significantly boosted both the performance and robustness of pattern recognition systems. In semantic segmentation tasks, the model learns by identifying and interpreting objects and regions within a single input image. To enhance the performance of semantic segmentation, researchers often employ strategies such as expanding the receptive field and incorporating contextual relationships, enabling the model to better capture and represent image features [9], [10], [11].

Change detection, akin to semantic segmentation, is also a pixel-level recognition task. However, unlike semantic segmentation, which involves single-image inputs, change detection requires bi-temporal images to be fed into the network during training. This setup enables the model to identify and learn regions that have undergone changes over time. Consequently, researchers have emphasized that change detection demands not only a robust feature fitting mechanism for individual images but also the construction of interaction mechanisms between features extracted from bi-temporal images [12], [13], [14]. These interaction mechanisms enhance the model's ability to efficiently learn and detect changes, thereby improving the overall performance of change detection systems [15].

Regarding feature interaction mechanisms, many researchers have proposed various solutions. Some researchers utilize attention mechanisms to construct feature interactions [15], [16], [17], [18], while others employ conventional convolutional modules for this purpose [19], [20]. Furthermore, in the Changer model, Fang et al. systematically analyzed multiple approaches for feature interaction, including aggregation-distribution, channel exchange, spatial exchange, and flow-based dual-alignment fusion [21].

Unlike previous studies, we propose a novel differential feature learning mechanism to construct the interaction between bi-temporal features. As shown in Fig. 1, we use cosine similarity to calculate the spatial and channel similarities of common bi-temporal images in change detection. Through visualization and quantitative analysis, we discovered that bi-temporal images can not only compute spatial differences to represent differential features but also calculate channel differences to express differential features. However, traditional differential feature learning primarily focuses on computing spatial differences [12], [22], [23]. In contrast, our approach introduces a channel-based computation of global differences across feature maps. By integrating spatial difference information with channel difference information, we design a Channel-Spatial Difference Weighting (CSDW) module based on the cosine similarity algorithm. This module enables the bi-temporal feature maps to more effectively focus on change regions. Our key contributions are summarized as follows:

(1) By computing spatial and channel information differences of bitemporal features, we designed a novel differential feature learning module specifically for change detection tasks.

(2) During the decoding stage, we developed a layer-exchange decoder (LED) that progressively enhances feature interactions through a layer-by-layer exchange mechanism.

1. **Decoder In Change Detection**
The change detection task is similar to the semantic segmentation task, as both typically adopt an encoder-decoder structure. The main goal of the encoding process is to extract features from bi-temporal images and generate a bi-temporal feature pyramid. This bi-temporal feature pyramid encapsulates multi-level information from the bi-temporal images. Generally, pixel-level prediction models fully utilize information at different levels to decode and restore the predicted mask. For the decoding part of the change detection task, researchers have explored various optimization strategies. Currently, the commonly used decoding methods for change detection include the following:

*1.1 Differential Feature Fusion + Layer-by-Layer Decoding*
The approach of differential feature fusion combined with layer-by-layer decoding involves first extracting features from bi-temporal images using a Siamese neural network. Then, differential features are calculated from the extracted bi-temporal features. Finally, these differential features are decoded to produce the prediction results [24], [25]. We argue that the calculation of differential features essentially disrupts the original bi-temporal image feature information, which is detrimental to the model's ability to effectively learn the complete information from bi-temporal images. Lin et al. proposed the EMAF model [26], which first constructs fused bi-temporal features using a

Foreground Module and then optimizes the decoding process for change target fitting by supervising the fused bi-temporal features with contour information. Zhu et al. proposed MDAFormer [27], which also uses a Siamese network with shared weights to extract a bi-temporal feature pyramid. Then, the Feature Difference Aggregation Module is used to perform differential feature fusion on the bi-temporal feature pyramid, resulting in a differential feature pyramid. Finally, a layer-by-layer decoding mechanism is employed to obtain the predicted mask.

*1.2 Bi-Temporal Feature Pyramid Decoding*

Tan et al. proposed the PRX-Change model [28], which uses a Siamese encoder in the encoding part to extract bi-temporal features. Meanwhile, a cross-attention mechanism is employed to perform aggregation-distribution computations on the bi-temporal features. In the decoding part, a simple layer-by-layer fusion decoder is used to decode the aggregated-distributed bi-temporal features. Feng et al. introduced the DMINet model [22], which calculates the difference and performs channel concatenation for multi-level bi-temporal features during the decoding stage to generate multi-level difference feature maps. These difference features are then progressively fused through a hierarchical Aggregator module, with partial predictions performed at various scales. Zhao et al. proposed the SGSLN model [14], which constructs a bi-temporal feature pyramid in the decoding phase by employing a channel-swapping strategy for bi-temporal features. Additionally, a three-branch layer-by-layer decoding structure is used to generate the predicted masks. Dong et al. proposed the EfficientCD model [29], which first uses EfficientNet as a Siamese network to extract bi-temporal features. Then, the ChangeFPN architecture is constructed to obtain a bi-temporal differential feature pyramid. Finally, a layer-by-layer decoding module is designed for the bi-temporal feature pyramid to perform decoding computations.

**2. Feature Interaction In Change Detection**
In change detection, feature interaction serves a dual purpose: calculating differences between bi-temporal features and facilitating information exchange between bi-temporal images. Since bi-temporal images are geographically co-located, there is a strong correlation between them, making such information exchange both practical and effective. Lin et al. propose a token exchange-based difference evaluation method [30]. This method involves exchanging tokens of bi-temporal images, followed by the application of a multi-head attention mechanism to highlight and model the differences between these tokens. By exchanging image information within bi-temporal images, this approach enhances information supplementation between the two images.

Noman et al. [13] introduced the Change-Enhanced Features Fusion Module (CEFF). CEFF performs channel-level weighted fusion on bi-temporal feature maps, optimizing the effectiveness of feature interaction. Its core innovation lies in adjusting the weights of each channel for bi-temporal features, allowing it to highlight feature channels with significant semantic changes while suppressing those that may contain noise.

Wei et al. proposed the Temporospatial Interactive Attention Module (TIAM), a feature interaction module designed to process bi-temporal feature maps, addressing interference caused by geometric perspective differences and temporal style variations in change detection tasks [31]. The core concept of TIAM is to construct Gram matrices between bi-temporal features to calculate spatiotemporal attention scores, capturing temporal and spatial correlations. To achieve this, TIAM first normalizes and applies convolution operations to the features, ensuring appropriate scales and dimensions. It then calculates the spatial and temporal matching relationships between bi-temporal features using similarity matrices. Finally, TIAM integrates these matching relationships into the feature maps and fuses the bi-temporal features through weighted methods, generating enhanced features that include spatiotemporal correlations.

Since change detection tasks focus on identifying changes between bi-temporal images rather than recognizing specific semantic categories, the features of bi-temporal images can be exchanged to enhance information interaction. The feature exchange mechanism promotes information flow between bi-temporal features, strengthening the representation capability of change detection models [14], [21], [29], [32].

In this paper, through investigating various bi-temporal feature interaction mechanisms, we found that most existing methods are limited to the spatial dimension, focusing on interactions between corresponding pixel feature vectors of bi-temporal features. By analyzing the information contained in bi-temporal features, we observed that the differences between bi-temporal features can also be reflected in the channel dimension. Based on this observation, we designed a Channel-Spatial Difference Weighting (CSDW) module that combines channel and spatial information to serve as the interaction mechanism for bi-temporal features.

# METHODS
## 1. Overall Architecture
IIn this paper, we optimize the change detection task from two perspectives. Firstly, in change detection tasks, the optimization of feature interaction methods can enhance the model's ability to perceive differential features. We designed a module that combines channel and spatial dimensions to compute differential features. Additionally, during the decoding stage, we constructed a decoding structure based on the Layer-Exchange method to strengthen the interaction of bi-temporal features. Therefore, we reinforced the interaction of bi-temporal features in both the encoding and decoding stages of the change detection model. As illustrated in Figure 2, the structure of our method is as follows.

Firstly, we employ Swin Transformer V2 [33] (SwinTV2) as the backbone network to leverage its powerful global information learning capabilities, as shown in Figure 2. To establish a bi-temporal feature interaction mechanism, we propose the Channel-Spatial Difference Weighting (CSDW) module to enhance the model's sensitivity to differential features. Second, we adopt the ChangeFPN module [29] to process the bi-temporal feature pyramid. At this stage, the bi-temporal feature pyramid undergoes further interaction through a layer-exchange mechanism. Finally, in the decoding stage, we design a simple feature fusion module based on the layer-exchange mechanism, named Layer-Exchange Decoder (LED), to process the bi-temporal feature pyramid and generate the final prediction output. On the right side of the "overall architecture" in Figure 2, the feature maps highlighted with green boxes are concatenated to compute auxiliary losses, with a weight of 0.3. The feature maps highlighted with blue boxes are concatenated to compute the primary loss and are also used as the output for inference. All loss functions employed are cross-entropy losses.

## 2. Channel-Spatial Difference Weighting
In change detection models, Siamese neural networks are commonly used to encode bi-temporal images, thereby generating Siamese feature pyramids. Throughout the encoding process, we perform aggregation and distribution operations on the bi-temporal features extracted at various levels of SwinTV2. Using this approach, the model calculates differential weights based on the bi-temporal features during the Siamese encoding process and applies layer-by-layer weighting to the bi-temporal features in the encoding stage. This enhances the sensitivity of the change detection model to change features during the encoding phase.

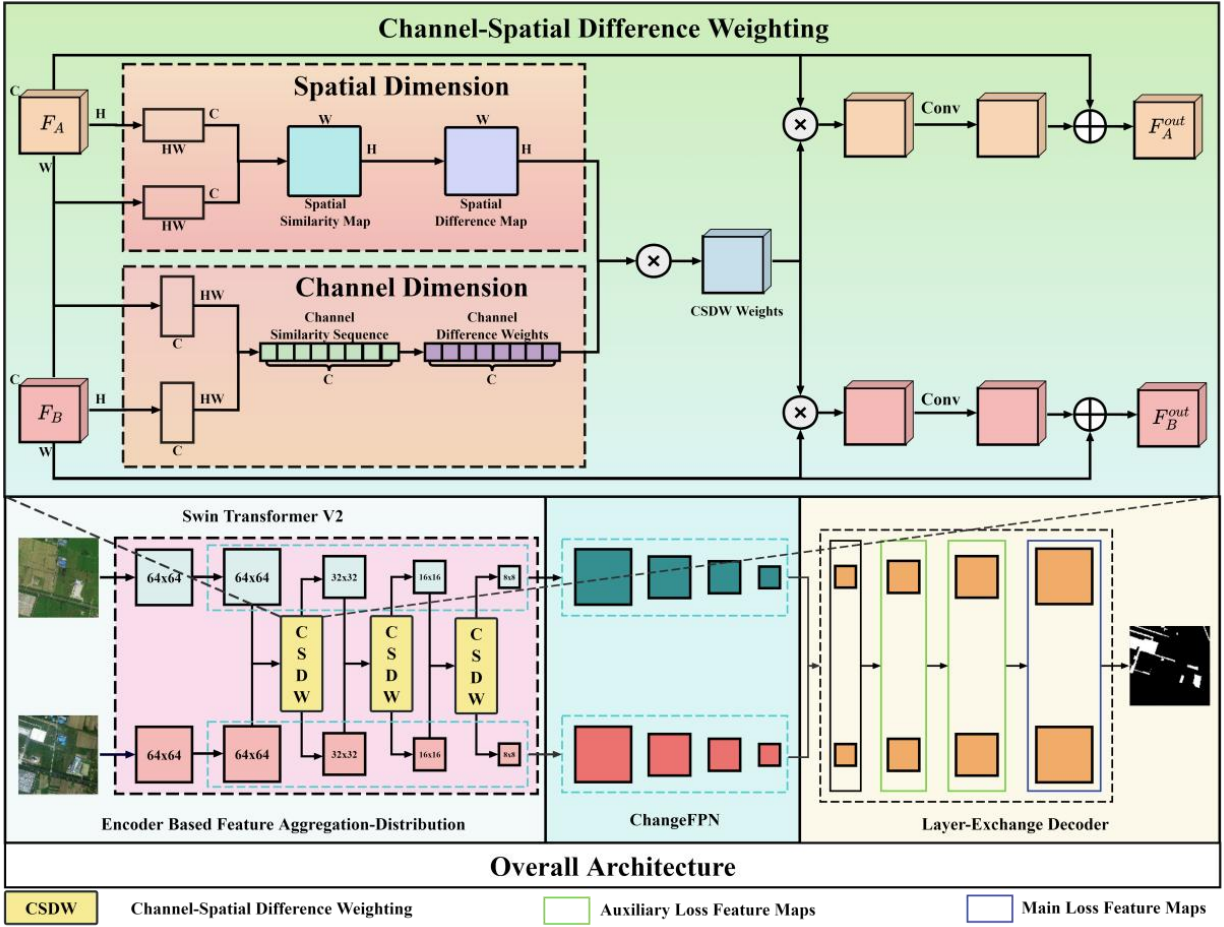

*Fig. 2. LENet Overall Architecture*

In computer vision tasks, feature extraction is typically applied to image data to construct high-dimensional feature map matrices. Features in the channel dimension are primarily generated through weighted computations using multiple convolutional kernels applied to the input feature maps. Since the convolutional kernels are initialized with different parameters, the features in each channel inherently focus on distinct aspects of the input, potentially representing edges, textures, shapes, or more abstract patterns. Spatial features, on the other hand, emphasize the relationships and layouts of pixels or regions within the image. Spatial features reflect the structural characteristics of objects, aiding in the identification and analysis of their shape, size, and layout. Therefore, we believe that for feature maps, they not only possess strong representational capabilities in spatial dimension for the original images but also exhibit abundant information representation channel-wise. Therefore, in change detection tasks, bi-temporal feature maps exhibit differences not only in the spatial dimension but also in the channel dimension.

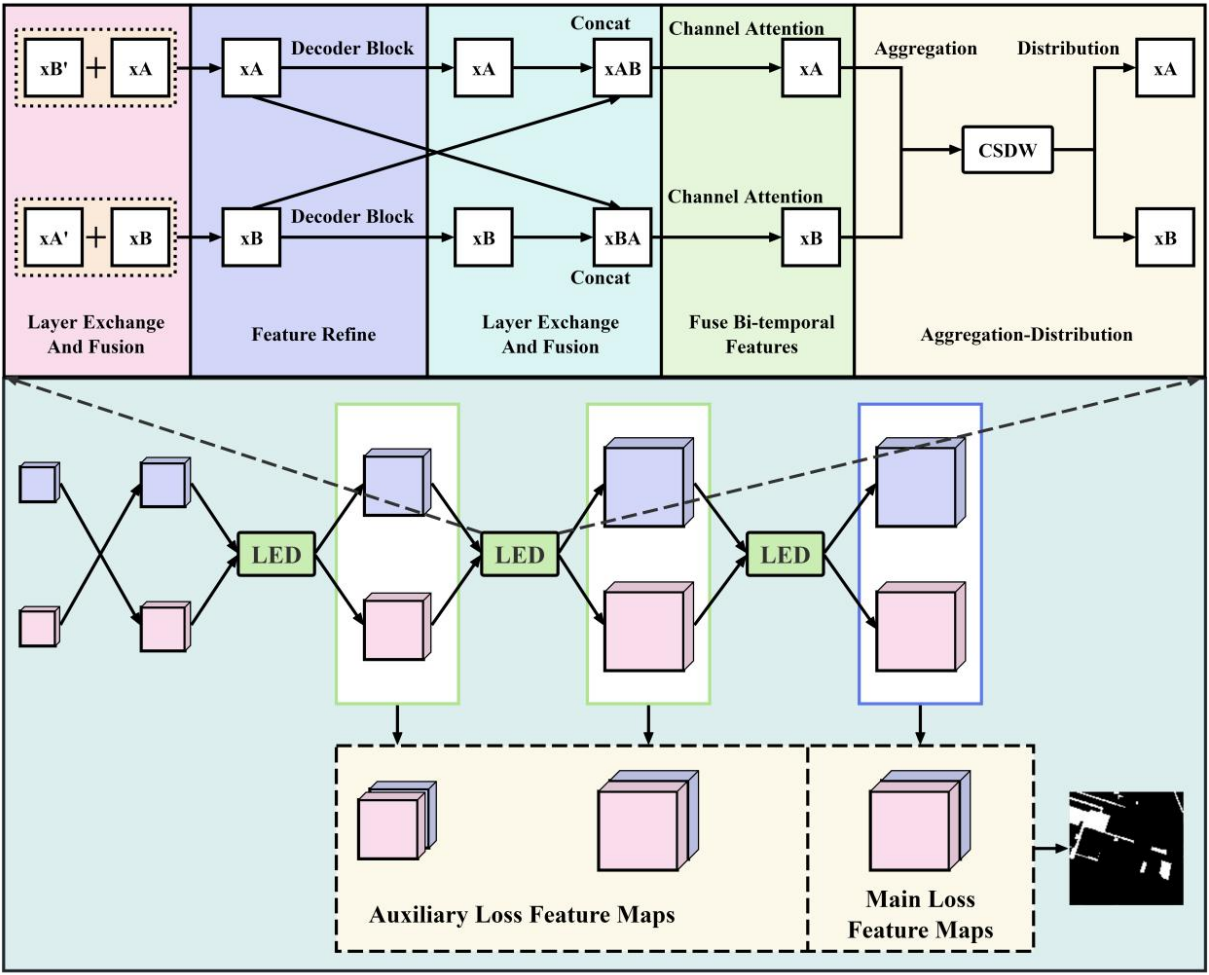

*Fig. 3. Layer-Exchange Decoder In LENet*

In this study, we propose the Channel-Spatial Difference Weighting (CSDW) module to learn the differences in bi-temporal feature maps across the spatial and channel dimensions, as shown in Figure 2 above. Overall, by utilizing cosine similarity to calculate differential features, the CSDW module applies differential weights to bi-temporal feature maps in both the spatial and channel dimensions. This weighting approach enhances the sensitivity of bi-temporal feature maps to change features.

The CSDW module generates change weights by calculating the cosine similarity between the feature maps of the two images and applying these weights to the feature maps, thereby achieving weighted processing of feature differences. The calculation method for the CSDW module is as follows:

$$F_c^{(i)} = \mathcal{R}(\,permute\,(F_i,(0,2,3,1)),(-1,C)), \; i \in \{A,B\} \tag{1}$$

$$\phi_c = \mathcal{R}\left(\frac{F_c^{(A)} \cdot F_c^{(B)}}{\|F_c^{(A)}\| \cdot \|F_c^{(B)}\|}, (N,H,W)\right) \tag{2}$$

$$W_c = 1 - \sigma(\,unsqueeze\,(\phi_c,1)) \tag{3}$$

$$F_s^{(i)} = \mathcal{R}(F_i,(N,C,-1)), \; i \in \{A,B\} \tag{4}$$

$$\phi_s = \mathcal{R}\left(\frac{F_s^{(A)} \cdot F_s^{(B)}}{\|F_s^{(A)}\| \cdot \|F_s^{(B)}\|}, (N, C)\right) \tag{5}$$

$$W_s = 1 - \sigma(unsqueeze(\phi_s, 1, 1)) \tag{6}$$

$$W = W_c \times W_s \tag{7}$$

$$F_i^{out} = Conv_i(W \times F_i) + F_i, \ i \in \{A, B\} \tag{8}$$

Among them, $F_A$ and $F_B$ represent the input feature maps with dimensions $(N, C, H, W)$. $\mathcal{R}$ represents reshape function to change the dimension of the input. $\phi$ represents the cosine similarity result, calculated as the dot product of the flattened feature maps divided by the product of their norms. $\sigma$ denotes the sigmoid function. W represents the change weights. Conv denotes the convolutional blocks, which further process the input feature maps. The final output feature maps $F_A^{out}$ and $F_B^{out}$ are the results of adding the residual module outputs to the original input feature maps. Through this series of operations, the model can effectively capture and process the change information of bi-temporal images, enhancing sensitivity and understanding of the change regions.

Besides, we use ChangeFPN [29] to process the bi-temporal feature pyramid extracted in the encoding stage. Thus, the Siamese-encoder not only focuses on the features of a single temporal image but also comprehensively considers the features of both images, enhancing the robustness of feature representation.

### 3. Layer-Exchange Decoder

In change detection tasks, the goal of the decoding stage is to generate high-quality change detection maps based on the feature information extracted during the encoding stage. Due to the significant scale variations of objects in remote sensing images, feature pyramids are typically constructed during the encoding stage to enhance the model's representation capability for targets at different scales. Consequently, during the decoding stage, the model can utilize multi-scale feature information extracted in the encoding stage.

Furthermore, we observed that bi-temporal images, being from the same geographic location, exhibit strong correlations between their features. To establish these correlations in change detection models, we introduced a layer-exchange feature fusion mechanism during the layer-by-layer decoding process to facilitate the learning of correlations between bi-temporal features. In the progressive decoding stage, we employed the SwinTV2Block module to optimize the features, enhancing the model's fitting capability.

As shown in Figure 3, we provide a detailed explanation of the Layer-Exchange Decoder (LED) in LENet. The structure of the LED is illustrated in the upper part of Figure 3. Here, xA' and xB' are the bi-temporal features from the previous layer. Firstly, through the layer-exchange mechanism, the feature maps xA and xB from the two temporal images are cross-fused to generate new feature maps xA and xB. Secondly, these feature maps are further refined using the SwinTV2 block decoding block to enhance feature representation capabilities. Additionally, to further facilitate feature interaction, we performed residual feature fusion based on the layer-exchange mechanism. Subsequently, the feature maps are weighted using the channel attention mechanism to emphasize important features. Finally, the features are further weighted through the CSDW-based feature aggregation-distribution steps to enhance feature interaction. The processed feature maps are then concatenated and fed into the decoding head for generating change detection results.

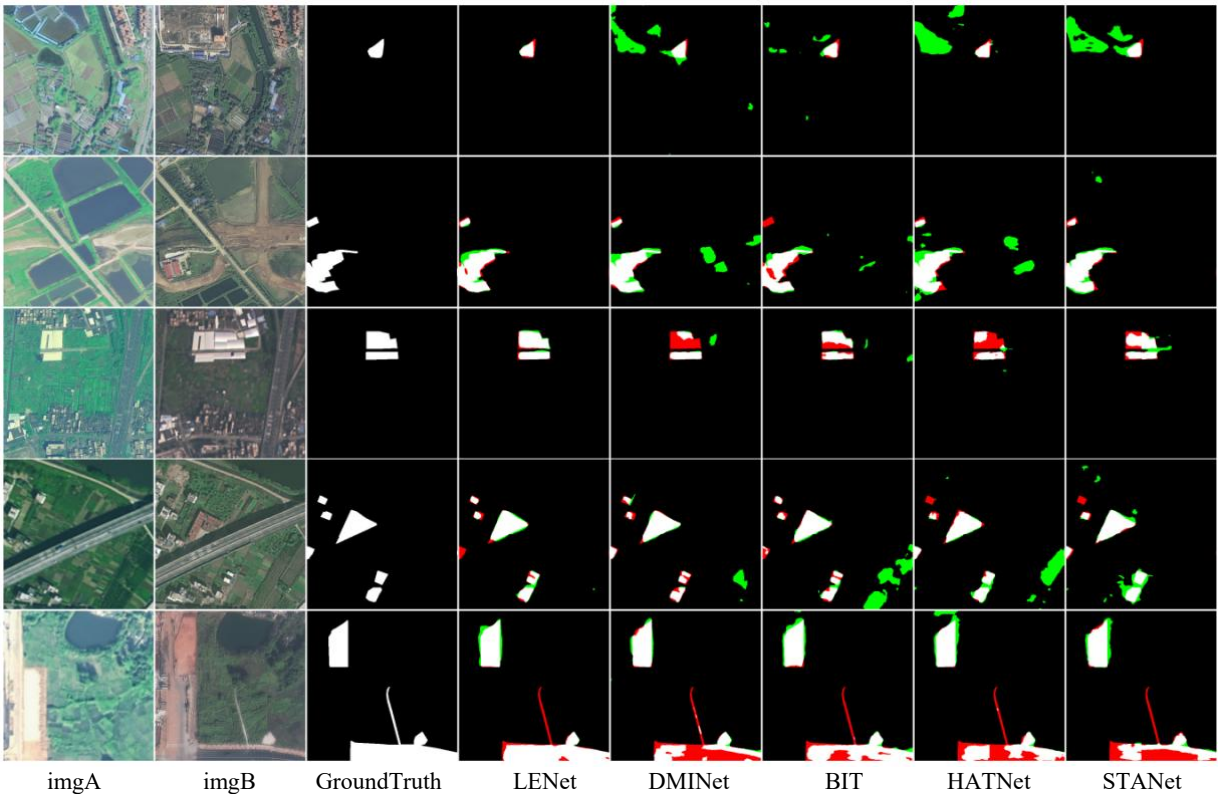

Fig. 4. *Visualization results in CLCD dataset*

TABLE I
QUANTITATIVE RESULTS ON THE CLCD DATASET

| Model | OA | IoU | F1 | Rec | Prec |
|---|---|---|---|---|---|
| ACABFNet [34] | 95.53 | 51.45 | 67.94 | 63.63 | 72.88 |
| STANet [35] | 95.50 | 51.49 | 67.97 | 64.16 | 72.26 |
| P2V [36] | 95.84 | 54.10 | 70.22 | 65.93 | 75.11 |
| MSCANet [37] | 96.05 | 55.83 | 71.65 | 67.07 | 76.91 |
| HATNet [38] | 96.09 | 56.90 | 72.53 | 69.42 | 75.94 |
| BIT [39] | 96.46 | 58.36 | 73.71 | 66.63 | 82.47 |
| DSIFN [20] | 96.55 | 59.42 | 74.54 | 67.86 | 82.69 |
| MIN-Net [40] | 96.56 | 62.08 | 76.60 | 75.70 | 77.53 |
| AMTNet [41] | -- | 62.35 | 76.81 | 75.06 | 78.64 |
| CGNet [42] | 96.82 | 62.67 | 77.05 | 71.71 | 83.25 |
| CACG-Net [43] | -- | 64.76 | 78.61 | 76.71 | 80.61 |
| EfficientCD [29] | 96.98 | 65.14 | 78.89 | 75.83 | 82.21 |
| LENet | 97.15 | 66.83 | 80.12 | 77.09 | 83.39 |

Among these, the concatenated feature maps within the green box are used to calculate auxiliary loss, while the concatenated feature maps within the blue box are used to calculate the main loss and serve as the output for prediction results. Throughout the entire

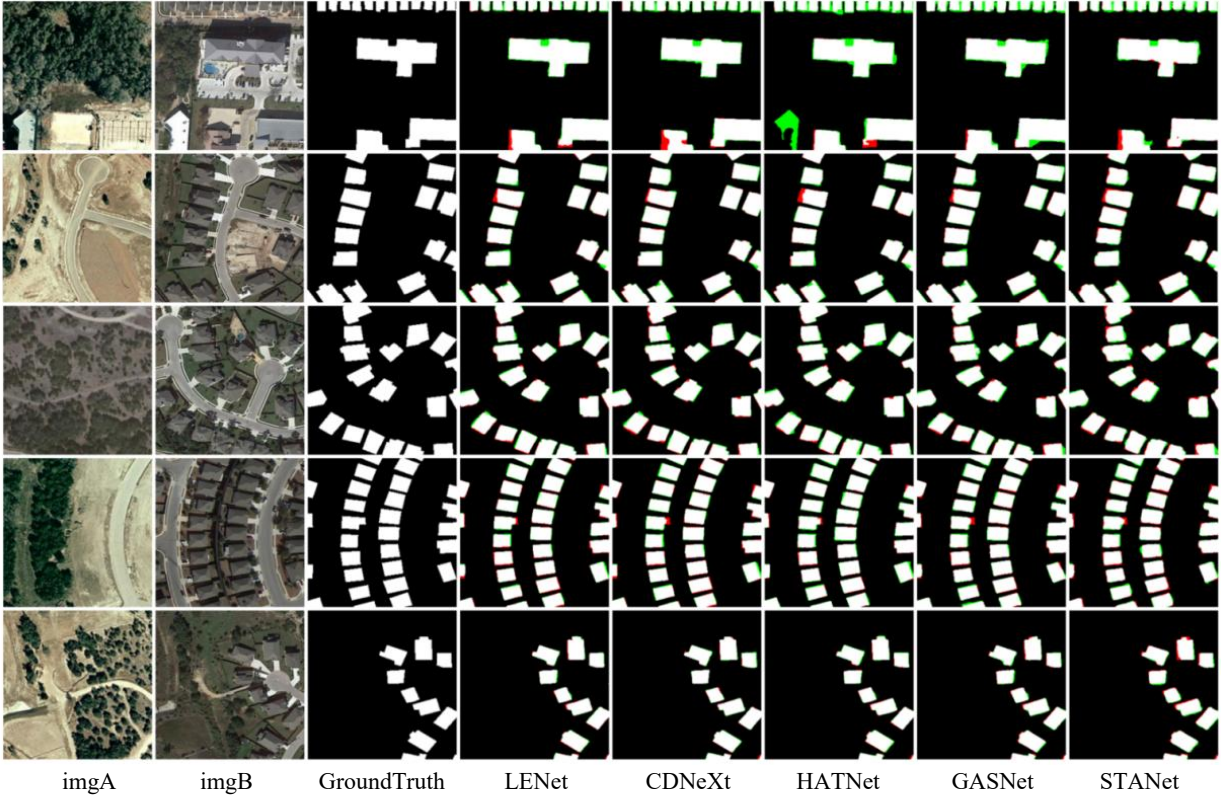

| imgA | imgB | GroundTruth | LENet | CDNeXt | HATNet | GASNet | STANet |

Fig. 5. *Visualization results in LEVIR-CD dataset*

Layer-Exchange Decoder process, multi-level feature fusion and exchange mechanisms are employed to progressively optimize and enhance the features of bi-temporal images, thereby improving the accuracy and robustness of change detection.

TABLE II
QUANTITATIVE RESULTS ON THE LEVIR-CD DATASET

| Model | OA | IoU | F1 | Rec | Prec |
|---|---|---|---|---|---|
| STANet [35] | 99.02 | 81.85 | 90.02 | 87.13 | 93.10 |
| ChangeFormer [44] | 99.04 | 82.66 | 90.50 | 90.18 | 90.83 |
| ACABFNet [34] | -- | -- | 90.68 | 89.96 | 91.40 |
| Changer [21] | -- | -- | 92.06 | 90.56 | 93.61 |
| DMATNet [23] | 98.25 | 84.13 | 90.75 | 89.98 | 91.56 |
| GASNet [45] | 99.11 | -- | 91.21 | 90.62 | 91.82 |
| ACAHNet [46] | 99.14 | 84.35 | 91.51 | 90.68 | 92.36 |
| HATNet [38] | -- | 84.41 | 91.55 | 90.23 | 92.90 |
| PCAANet [47] | 98.26 | 85.22 | 92.02 | 90.67 | 93.41 |
| EfficientCD [29] | 99.22 | 85.55 | 92.21 | 91.22 | 93.23 |
| CACG-Net [29] | -- | 85.68 | 92.29 | 92.41 | 92.16 |
| CDNeXt [31] | 99.24 | 85.86 | 92.39 | 90.92 | 93.91 |
| RSBuilding [48] | -- | 86.19 | 92.59 | 91.80 | 93.39 |
| LENet | 99.26 | 86.30 | 92.64 | 91.22 | 94.12 |

**RESULTS**

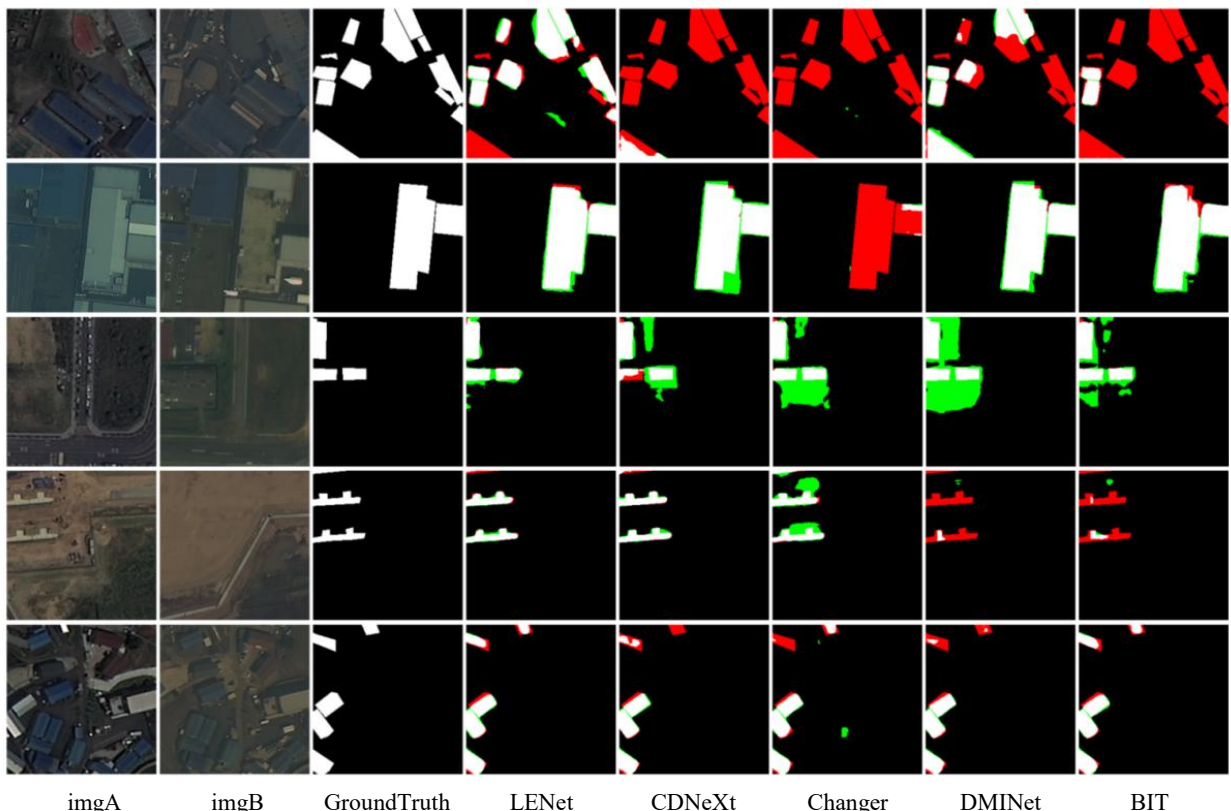

imgA　　imgB　　GroundTruth　　LENet　　CDNeXt　　Changer　　DMINet　　BIT

*Fig. 6. Visualization results in S2Looking dataset*

**1. Datasets**

In this study, we leveraged four sub-meter resolution datasets—LEVIR-CD [35], PX-CLCD [49], S2Looking [50], and CLCD [37] —to showcase the robustness and versatility of our change detection algorithm across diverse environments and scenarios. The LEVIR-CD dataset is a crucial resource for building change detection. It contains 637 pairs of high-resolution images (0.5 m/pixel, 1024×1024 pixels) obtained from Google Earth, with 31,333 instances of building changes. The dataset is divided into 445 training pairs, 64 validation pairs, and 128 testing pairs, further segmented into 256×256 pixel patches with a 64-pixel overlap.

TABLE III
QUANTITATIVE RESULTS ON THE S2LOOKING DATASET

| Model | OA | IoU | F1 | Rec | Prec |
| --- | --- | --- | --- | --- | --- |
| BIT [39] | 99.24 | 47.94 | 64.81 | 58.15 | 73.20 |
| HATNet [38] | -- | 47.08 | 64.02 | 60.90 | 67.48 |
| FHD [51] | -- | 47.33 | 64.25 | 56.71 | 74.09 |
| CGNet [42] | -- | 47.41 | 64.33 | 59.38 | 70.18 |
| SAM-CD [52] | -- | 48.29 | 65.13 | 58.92 | 72.80 |
| DMINet [53] | 99.20 | 48.33 | 65.16 | 62.13 | 68.51 |
| PCAANet [47] | 99.22 | 48.54 | 65.36 | 61.54 | 69.68 |
| CDNeXt [31] | -- | 50.05 | 66.71 | 63.08 | 70.78 |
| Changer [21] | 99.26 | 50.47 | 67.08 | 62.04 | 73.01 |
| LENet | 99.29 | 51.19 | 67.71 | 61.90 | 74.72 |

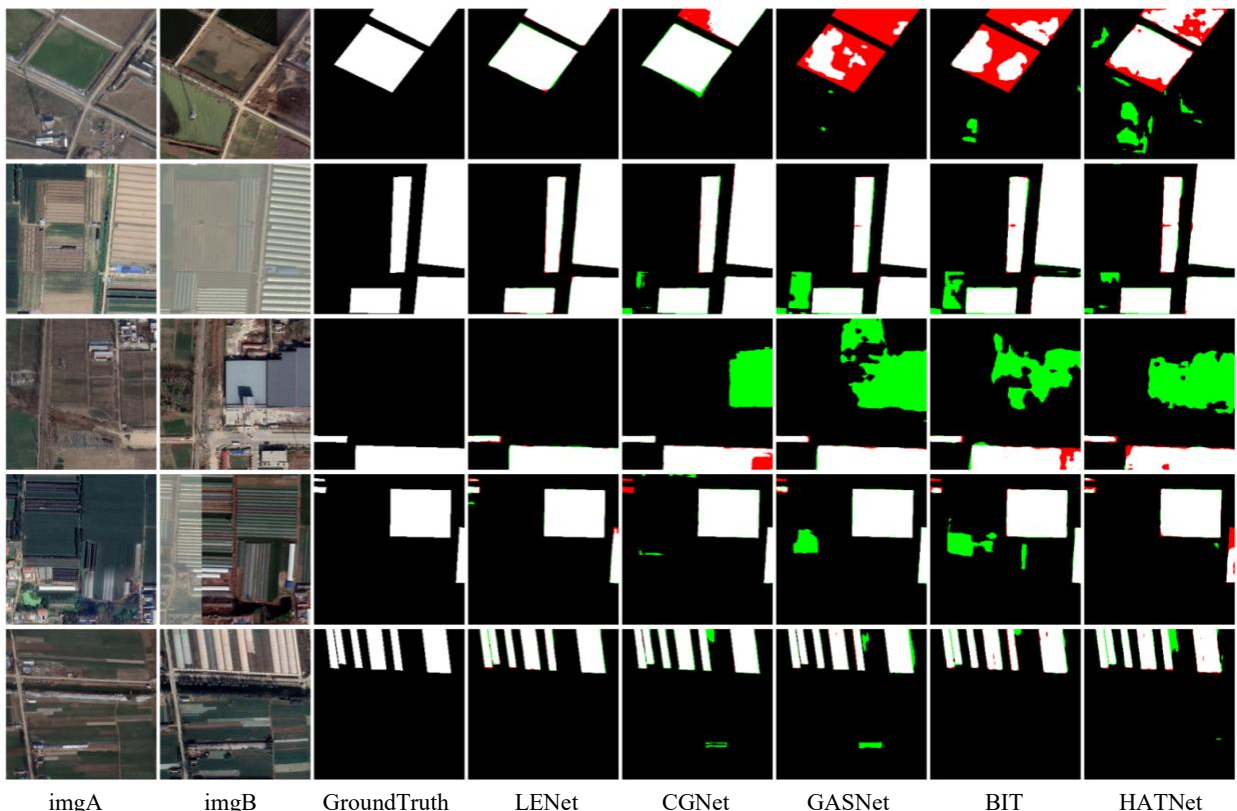

Fig. 7. *Visualization results in PX-CLCD dataset*

The PX-CLCD dataset focuses on cultivated land change detection and consists of 5170 pairs of 1-meter spatial resolution bi-temporal images (256×256 pixels). The dataset is split into training, validation, and testing sets in a 6:2:2 ratio, representing cultivated land changes between 2018 and 2021. The S2Looking dataset comprises 5000 image pairs (1024×1024 pixels) with over 65,920 annotated changes derived from rural satellite images (0.5–0.8 m/pixel). It is divided into training, evaluation, and testing sets in a 7:1:2 ratio. The CLCD dataset contains 600 pairs of farmland change detection images (512×512 pixels, 0.5–2 m resolution) collected by Gaofen-2 in Guangdong, China, during 2017 and 2019. The dataset includes 320 training pairs, 120 validation pairs, and 120 testing pairs. For training, random 256×256 patches are used, while sliding window predictions are employed for inference.

TABLE IV
QUANTITATIVE RESULTS ON THE PX-CLCD DATASET

| Model | OA | IoU | F1 | Rec | Prec |
|---|---|---|---|---|---|
| HATNet [38] | 98.50 | 88.99 | 94.18 | 93.83 | 94.53 |
| MSCANet [37] | 98.50 | 89.00 | 94.18 | 93.95 | 94.41 |
| BIT [39] | 98.76 | 90.78 | 95.17 | 94.80 | 95.54 |
| GASNet [45] | 98.99 | 92.51 | 96.11 | 96.42 | 95.80 |
| DMINet [53] | 99.04 | 92.83 | 96.28 | 96.31 | 96.25 |
| SNUNet3+ [49] | 99.19 | 93.61 | 96.64 | 96.79 | 96.60 |
| CGNet [42] | 99.17 | 93.82 | 96.81 | 97.33 | 96.30 |
| LENet | 99.32 | 94.86 | 97.36 | 97.08 | 97.65 |

**2. Implementation Detail**

We trained the LENet model on the Nvidia A100 GPU. And, we used three methods: RandomRotate, RandomFlip and PhotoMetricDistortion for data enhancement. In terms of model optimization, the AdamW optimizer was utilized. Throughout the experimental stage, we continuously monitored the IoU metric on the validation set, earmarking the best-performing model for subsequent final evaluation.

### 3. Evaluation Metrics

We used these metrics to evaluate our model like precision (Prec), recall (Rec), overall accuracy (OA), F1-score (F1), and Intersection over Union (IoU). Its calculation formula is as follows:

$$\text{IoU} = \frac{\text{TP}}{\text{TP} + \text{FN} + \text{FP}} \tag{9}$$

$$\text{Prec} = \frac{\text{TP}}{\text{TP} + \text{FP}} \tag{10}$$

$$\text{Rec} = \frac{\text{TP}}{\text{TP} + \text{FN}} \tag{11}$$

$$F1 = 2\frac{P \cdot R}{P + R} \tag{12}$$

$$\text{OA} = \frac{\text{TP} + \text{TN}}{\text{TP} + \text{TN} + \text{FN} + \text{FP}} \tag{13}$$

### 4. Quantify analysis and visualize results with compared methods

In the field of remote sensing, change detection tasks have a wide range of applications, among which two common scenarios are monitoring the non-agriculturalization of cultivated land and illegal building detection. Specifically, LEVIR-CD and S2Looking datasets are designed for building change detection tasks, while CLCD and PX-CLCD are targeted at detecting changes in cultivated land. Based on these four datasets, we compared numerous state-of-the-art algorithms and conducted comprehensive experiments. As shown in Tables I through IV, the results indicate that LENet consistently outperforms its competitors across all major evaluation metrics, demonstrating its exceptional performance in change detection tasks.

Through an investigation of the corresponding datasets, we selected advanced models from recent years with different research focuses as comparative models for the experiments. These models cover cutting-edge research areas such as differential feature computation [53], integration with AI foundational models [52], utilization of massive datasets [54], attention mechanisms [41], [46], and other advanced topics in remote sensing change detection. By comparing with these advanced models, we aim to demonstrate the sufficient advantages of the proposed LENet in this paper.

To ensure the fairness of the experiments, we retrained certain comparative models for which the original papers did not provide results. Since these tasks involve binary change detection, we selected Intersection-over-Union (IoU) for the foreground change class as the primary evaluation metric, alongside other metrics such as F1-score, Recall, Precision, and Overall Accuracy to assess the model's overall performance comprehensively.

The experimental results on the CLCD, LEVIR-CD, PX-CLCD, and S2Looking datasets demonstrate that LENet achieves outstanding performance across multiple key evaluation metrics. Taking the CLCD and LEVIR-CD datasets as examples, LENet surpasses existing methods in IoU, F1, Recall, and Precision by varying degrees. On the CLCD dataset, LENet achieves an IoU of 66.83% and an F1 of 80.12%, outperforming representative methods such as EfficientCD (IoU 65.14%, F1 78.89%). Meanwhile, on the LEVIR-CD dataset, LENet exhibits comprehensive superiority in IoU, F1, Recall, and Precision, demonstrating its robust and efficient detection capabilities and further confirming its advantages in handling urban and large-scale scene change detection.

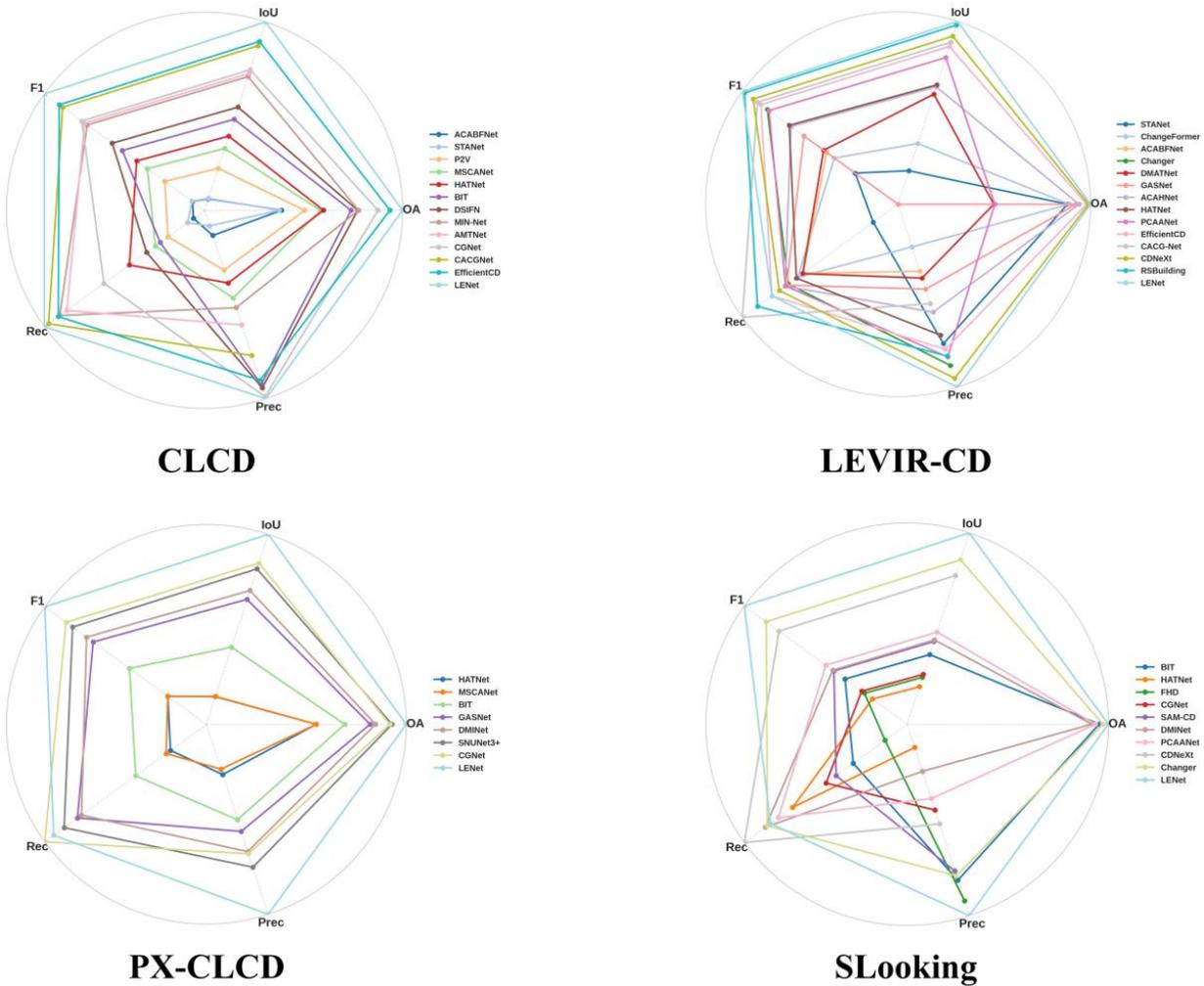

*Fig. 8. Comparative Radar Chart of LENet and Other Models Across Multiple Datasets*

LENet continues to deliver impressive results on the more challenging PX-CLCD and S2Looking datasets. On PX-CLCD, LENet sets new benchmarks in IoU, F1, Recall, and Precision, surpassing previous best-performing methods. On the S2Looking dataset, LENet achieves an IoU of 51.19% and an F1 score of 67.71%, further solidifying its leading position in remote sensing change detection. Overall, LENet demonstrates consistent and superior performance across diverse remote sensing datasets of varying difficulty, proving its effectiveness in feature extraction, difference representation, and fine-grained object segmentation.

Additionally, Figures 4 through 7 provide visualizations of LENet's test results on the CLCD, LEVIR-CD, S2Looking and PX-CLCD datasets. In these visualizations, True Positives (TP) are represented by white pixels, True Negatives (TN) by black pixels, False Positives (FP) by green pixels, and False Negatives (FN) by red pixels. Comparing these visual results reveals that LENet performs exceptionally well across different datasets and diverse application scenarios, aligning closely with the ground truth annotations. Its detection results not only exhibit high accuracy but also demonstrate remarkable reliability. This further validates the practicality and robustness of LENet in remote sensing image change detection tasks.

Meanwhile, to provide a more intuitive illustration of LENet's superiority on these four datasets, we plotted radar charts of the comparison results, as shown in Figure 8. From these four radar charts, one can directly observe each method's overall performance on different datasets and evaluation metrics. A larger "spider web" area indicates a more balanced and advantageous performance in IoU, F1, Recall, Precision, and OA. As seen in all four datasets, LENet shows significant outward extensions in multiple metrics, highlighting its strengths in accuracy, completeness, and overall detection effectiveness. Compared with other methods, LENet exhibits a notable advantage in generalization and stability.

TABLE V
ABLATION STUDY IN IOU INDEX

| Model | Encoder (CSDW) | Decoder (LED) | LEVIR-CD | PX-CLCD | CLCD | S2Looking |
|---|---|---|---|---|---|---|
| LENet | × | × | 84.85 | 93.74 | 59.96 | 49.12 |
|  | √ | × | 85.53 | 94.34 | 61.03 | 50.05 |
|  | × | √ | 86.08 | 94.40 | 61.74 | 50.76 |
|  | √ | √ | 86.30 | 94.86 | 66.83 | 51.19 |

**5. Ablation study**

The ablation study focused on the Intersection over Union (IoU) index across four datasets—CLCD, LEVIR-CD, PX-CLCD, and S2Looking—reveals the significant impact of incorporating the CSDW module in the encoding stage and the LED into the LENet. In TABLE VI, the Encoder (CSDW) means that we used the CSDM as the aggregation-distribution module in the encoding stage. The Decoder (LED) means the we used the layer-exchange based decoder (LED) in the decoding stage. Furthermore, we used the normal encoder based the SwinTV2 and a normal decoder upsample layer-by-layer as the baseline.

The results in TABLE V indicate that both the Encoder (CSDW) and the LED significantly contribute to improving the IoU scores across all datasets. When neither the Encoder (CSDW) nor the LED is integrated (baseline), the performance is lower across all datasets. Specifically, the LEVIR-CD dataset achieves an IoU of 84.85, PX-CLCD achieves 93.74, CLCD achieves 59.96, and S2Looking achieves 49.12.

By adding the Encoder (CSDW) in the encoding stage, the IoU scores show a notable increase. For example, the IoU score improves from 84.85 to 85.53 for the LEVIR-CD dataset and from 93.74 to 94.34 for PX-CLCD. Similarly, the CLCD dataset sees an increase from 59.96 to 61.03, while the S2Looking dataset improves from 49.12 to 50.05. On the other hand, when the LED is included without the Encoder (CSDW), the IoU scores also improve across datasets. For instance, the LEVIR-CD dataset rises to 86.08, PX-CLCD increases to 94.40, CLCD improves to 61.74, and S2Looking reaches 50.76. This demonstrates the effectiveness of the LED in the decoding stage for enhancing feature interaction and fusion.

The integration of both the Encoder (CSDW) and the LED further boosts the IoU scores, showcasing their complementary roles. The LEVIR-CD dataset achieves the highest score of 86.30, PX-CLCD reaches 94.86, CLCD improves to 66.83, and S2Looking achieves 51.19. These results highlight that combining the CSDW module with the LED leads to the most comprehensive improvement, underscoring their synergistic effect in improving the representation and fusion of bi-temporal features in change detection tasks.

## DISCUSSION

### 1. Feature Interaction in Change Detection

In change detection tasks, feature interaction is a critical factor in enhancing model performance. Feature interaction enables thorough information exchange between bi-temporal images, thereby improving the model's ability to represent bi-temporal data. Through feature interaction mechanisms, the model's sensitivity to differential regions is enhanced, promoting the fusion and information sharing of bi-temporal features. Additionally, the layer-exchange mechanism only swaps bi-temporal image features at appropriate positions without altering the model structure, thus facilitating bi-temporal feature interaction without adding computational burden. Specifically, the Channel-Spatial Difference Weighting (CSDW) module applies weighted processing to bi-temporal features, allowing the regions of interest in bi-temporal features to focus more on change regions, thereby constructing multi-level feature interaction during the encoding stage. In the decoding stage, we employ the layer-exchange mechanism to achieve cross-fusion of bi-temporal features, followed by CSDW-based feature weighting during the decoding process, further optimizing the feature representation of change regions.

### 2. Layer-Exchange Mechanism In Change Detection

In change detection tasks, the layer-exchange mechanism provides an innovative solution for bi-temporal feature interaction. Since bi-temporal images are derived from the same geographic location, their features exhibit high correlation. By employing the layer-exchange mechanism during the decoding stage, we achieve cross-temporal interaction of bi-temporal features, enhancing the representation capability of change features. Compared to traditional feature fusion methods, the layer-exchange mechanism directly swaps bi-temporal feature layers, enabling deep-level information fusion while maintaining the structural simplicity and computational efficiency of the model. Through the exchange of corresponding feature layers, the layer-exchange mechanism allows the model to efficiently integrate bi-temporal information without increasing parameters, strengthening the information exchange between bi-temporal features and enhancing the representational capacity of the change detection model. Experimental results in this paper demonstrate that this layer-exchange decoding design significantly improves the model's performance in change detection tasks, achieving excellent results in both accuracy and robustness.

## CONCLUSION

In this study, we explored the computation of change information between bi-temporal images based on both spatial and channel dimensions, proposing the CSDW module to optimize the learning of differential features between bi-temporal features. Additionally, in the decoding stage, we designed a novel decoding module (LED) based on the layer-exchange mechanism to enhance the interaction of bi-temporal features during decoding.

Extensive experiments conducted on the CLCD, LEVIR-CD, PX-CLCD, and S2Looking datasets further validated the effectiveness of LENet. In future work, we will continue to explore the importance of feature exchange in change detection architectures, such as constructing change detection frameworks without any differential feature computation modules, and leveraging the feature exchange mechanism to investigate self-supervised learning methods in change detection tasks.